\documentclass[conference]{IEEEtran}
\IEEEoverridecommandlockouts    

\usepackage{tikz}

\newcommand\copyrighttext{%
  \footnotesize \textcopyright 2026 IEEE. Personal use of this material is permitted.
  Permission from IEEE must be obtained for all other uses, in any current or future
  media, including reprinting/republishing this material for advertising or promotional
  purposes, creating new collective works, for resale or redistribution to servers or
  lists, or reuse of any copyrighted component of this work in other works.}
\newcommand\copyrightnotice{%
\begin{tikzpicture}[remember picture,overlay]
\node[anchor=south,yshift=10pt] at (current page.south) 
  {\fbox{\parbox{\dimexpr\textwidth-\fboxsep-\fboxrule\relax}{\copyrighttext}}};
\end{tikzpicture}%
}

\usepackage{wrapfig}
\usepackage{amsmath,amsfonts}
\usepackage{algorithmic}
\usepackage{algorithm}
\usepackage{array}
\usepackage[caption=false,font=footnotesize,labelfont=rm,textfont=rm]{subfig}
\usepackage{textcomp}
\usepackage{stfloats}
\usepackage{url}
\usepackage{verbatim}
\usepackage{graphicx}
\usepackage{cite}
\usepackage[colorlinks,linkcolor=blue,citecolor=blue]{hyperref}
\usepackage{multirow}
\usepackage{diagbox}
\usepackage{makecell}
\usepackage{tabularray}
\usepackage{booktabs}

\title{\LARGE \bf A Style-Based Profiling Framework for Quantifying the Synthetic-to-Real Gap in Autonomous Driving Datasets}

\author{
	\parbox{\textwidth}{%
		\centering
Dingyi Yao, Xinyao Han, Ruibo Ming, Zhihang Song, Lihui Peng, Jianming Hu, Danya Yao and Yi Zhang%
	}%
	\thanks{Corresponding author: Lihui Peng. All authors are with the Department of
Automation, Tsinghua University, Beijing 100084, China (e-mail: \{ydy24, hanxy24, mrb22, song-zh22\}@mails.tsinghua.edu.cn; lihuipeng@tsinghua.edu.cn;  \{hujm, yaody, zhyi\}@mail.tsinghua.edu.cn)}%
}

\begin{document}
	
	\maketitle
	\thispagestyle{empty}
	\pagestyle{empty}

    \copyrightnotice
	\begin{abstract}
		Ensuring the reliability of autonomous driving perception systems requires extensive environment-based testing, yet real-world execution is often impractical. Synthetic datasets have therefore emerged as a promising alternative, offering advantages such as cost-effectiveness, bias free labeling, and controllable scenarios. However, the domain gap between synthetic and real-world datasets remains a major obstacle to model generalization. To address this challenge from a data-centric perspective, this paper introduces a profile extraction and discovery framework for characterizing the style profiles underlying both synthetic and real image datasets. We propose Style Embedding Distribution Discrepancy (SEDD) as a novel evaluation metric. Our framework combines Gram matrix-based style extraction with metric learning optimized for intra-class compactness and inter-class separation to extract style embeddings. Furthermore, we establish a benchmark using publicly available datasets. Extensive experiments demonstrate that SEDD aligns with human perception where traditional NR-IQA metrics fail. Specifically, it correctly identifies the superior fidelity of Virtual KITTI 2 over Virtual KITTI and quantifies the gap reduction achieved by  sim-to-real methods. This work provides a standardized proactive quality control paradigm that enables the systematic diagnosis of dataset deficiencies and guides the selection of optimal sim-to-real adaptation strategies, advancing future development of data-driven autonomous driving systems. 
	\end{abstract}
\section{Introduction}
\label{sec:intro}
 Ensuring the safety and reliability of autonomous driving perception systems requires extensive testing and validation. Recent research emphasizes environment-based testing, which evaluates system performance under diverse and dynamic conditions while revealing safety-critical risks \cite{fs}. However, conducting such testing in the real world is costly, time-consuming, and unsafe\cite{iv1,iv2}.

\begin{figure}[t]
  \centering
 \includegraphics[width=0.95\linewidth]{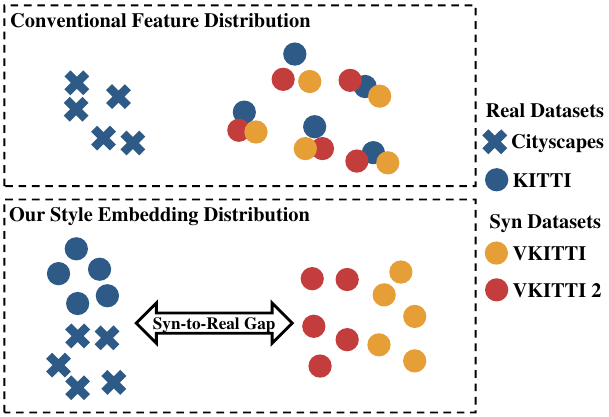}
   \caption{Schematic of our Style Embedding Distribution Discrepancy (SEDD). Conventional feature distributions are influenced by image content, separating different real datasets while clustering real datasets with their cloned synthetic counterparts. Our approach disentangles content and style, distinguishes between real and synthetic datasets, and measures the synthetic-to-real gap.}
   \label{fig:tu1}
\end{figure}

To overcome these limitations, synthetic datasets generated by autonomous driving simulation platforms have emerged as a promising supplementary data source \cite{song}. These synthetic datasets construct virtual environments using advanced graphics engines. For instance, the GTA-V\cite{GTAV} is derived from a commercial video game environment, whereas SYNTHIA \cite{SYNTHIA}, Virtual KITTI \cite{VKITTI1}, Virtual KITTI 2 \cite{VKITTI2} are developed using the Unity engine. Similarly, datasets such as SHIFT \cite{shift} are built upon the CARLA platform.

Compared to real-world datasets, synthetic datasets offer several notable advantages. First, they can be generated efficiently without the need for extensive time and labor-consuming data collection efforts. Second, annotations are programmatically generated with pixel-perfect precision, eliminating subjective bias that may arise in manual labeling processes. Additionally, synthetic environments enable full controllability over environmental variables, facilitating the generation of diverse scenarios. Finally, synthetic data can be targeted to generate rare scenarios critical to security verification that are difficult to capture in real world. Reflecting these advantages, prior studies have demonstrated their effectiveness in object detection and tracking \cite{iv3}, as well as lane detection and classification \cite{iv4}.

Despite the notable advantages of synthetic datasets, their application also presents new challenges, particularly in whether they support transfer to real-world scenarios. At the heart of this issue lies the synthetic-to-real gap, whose quantification is therefore crucial.

Existing methods for quantifying the synthetic-to-real gap in autonomous driving image datasets still have notable limitations. For example, common evaluation metrics include PSNR \cite{PSNR}, SSIM \cite{SSIM}, LPIPS \cite{LPIPS}, and FID \cite{FID}. These metrics assess quality according to the differences between generated images and their corresponding real reference images. However, synthetic datasets generated from simulation platforms typically provide only the final synthetic images without accompanying referenced real images, rendering these methods less applicable. Moreover, some studies \cite{29,31,gcv} indirectly evaluate dataset quality by comparing the performance differences on downstream tasks between real and synthetic datasets. Since the label spaces of different datasets may only partially overlap or even be entirely disjoint, these methods often have a limited scope of applicability.

To address these issues, we introduce a novel style-based profiling framework with a  metric, Style Embedding Distribution Discrepancy (SEDD). SEDD functions as a proactive diagnostic tool within the data generation pipeline. It allows researchers to screen environmental parameters to minimize the distributional distance to the real domain before large-scale rendering and aids in model selection for domain adaptation by identifying which translation method yields the most realistic style. The main contributions of this work are as follows: 
\begin{itemize}
\item 	 We innovatively model the discrepancies between synthetic and real data as differences in extracted style and propose a data fidelity metric SEDD. 

\item 	We utilize the publicly available datasets to establish a benchmark for evaluating the synthetic-to-real gap.

\item 	Experiment results demonstrate that SEDD outperforms traditional metrics by accurately ranking dataset fidelity.

\item We apply the proposed profiling framework to sim-to-real approaches, providing quantitative metrics and visualization tools that effectively validate the gap reduction achieved by various photorealism enhancement methods.
\end{itemize}

\section{Related Work}
\label{sec:relatedwork}

\subsection{NR-IQA for Single Images}
Since many synthetic datasets lack real-world counterparts \cite{shift,GTAV}, it is necessary to employ No-Reference Image Quality Assessment (NR-IQA) methods. Classical approaches such as BRISQUE \cite{brisque} rely on natural scene statistics with SVM regression. With the advent of deep learning, data-driven NR-IQA methods have have emerged, including NIMA \cite{nima}, which employs end-to-end CNN training, and Unique \cite{unique}, which leverages samples from multiple databases to generalize.

Although these methods have achieved commendable performance in generic contexts, they may incorrectly judge real-world datasets as having inferior quality. This error arises because conventional quality assessment techniques focus solely on the visual quality of an image (``how good the quality is"), neglecting its fidelity or similarity to real-world scenes (``how realistic the image is").


\subsection{Synthetic Datasets Generation and Enhancement for Autonomous Driving}
In the field of autonomous driving, synthetic datasets provide a new way to acquire data. One notable example is Virtual KITTI \cite{VKITTI1}. This dataset was developed by selecting five real-world video sequences from the KITTI \cite{KITTI} dataset as seed data, which were then cloned and modified for weather conditions in the Unity engine. Building on this foundation, Virtual KITTI 2 \cite{VKITTI2} was introduced as an enhanced version of Virtual KITTI. By taking advantage of new features available in the updated Unity engine, such as the high-definition rendering pipeline, Virtual KITTI 2 delivers improved visual effects with more realistic image details. 

After the rise of synthetic datasets, a number of data enhancement efforts have been proposed, working on the algorithms that improve the fidelity, such as color transfer \cite{colortransfer} based on statistical analysis, and CUT \cite{cut}, EPE \cite{EPE} based on image-to-image translation. In addition, CARLA2Real \cite{carla2real}, utilized photorealism enhancement technique to narrow the gap between CARLA platform and reality.  The target datasets for methods CARLA2KITTI and CARLA2CITY are KITTI \cite{KITTI} and Cityscapes \cite{Cityscapes}.


 Both generated and enhanced synthetic datasets need an objective metric to assess the synthetic-to-real gap.

\subsection{Synthetic-to-Real Gap Quantification for Autonomous Driving}
Gadipudi et al. \cite{gadipudi}  proposed a method that leverages feature embedding techniques to compute the Euclidean distance between real and synthetic datasets. However, its evaluation outcomes are susceptible to variations in image content and fail to disentangle content from style. This issue is particularly pronounced in scenarios where the Virtual KITTI dataset has a one-to-one correspondence with the KITTI dataset; in such cases, the computed feature distance may underestimate the stylistic differences between the two.
In addition, using unsupervised methods for dimensionality reduction before calculating distances will lead to significant information loss. 

Li et al. \cite{Li} evaluated datasets using a total of 11 metrics across three dimensions: formal quality, content quality, and utility quality. However, the fact that its indicators have full scores for all datasets on multiple indicators implies redundancy in these indicators, weakening the discriminative power of the methodology. Moreover, their handcrafted features may hinder the capture of high-level semantic differences.

\begin{figure*}[t]
    \centering
        \centering
        \includegraphics[width=0.95\textwidth]{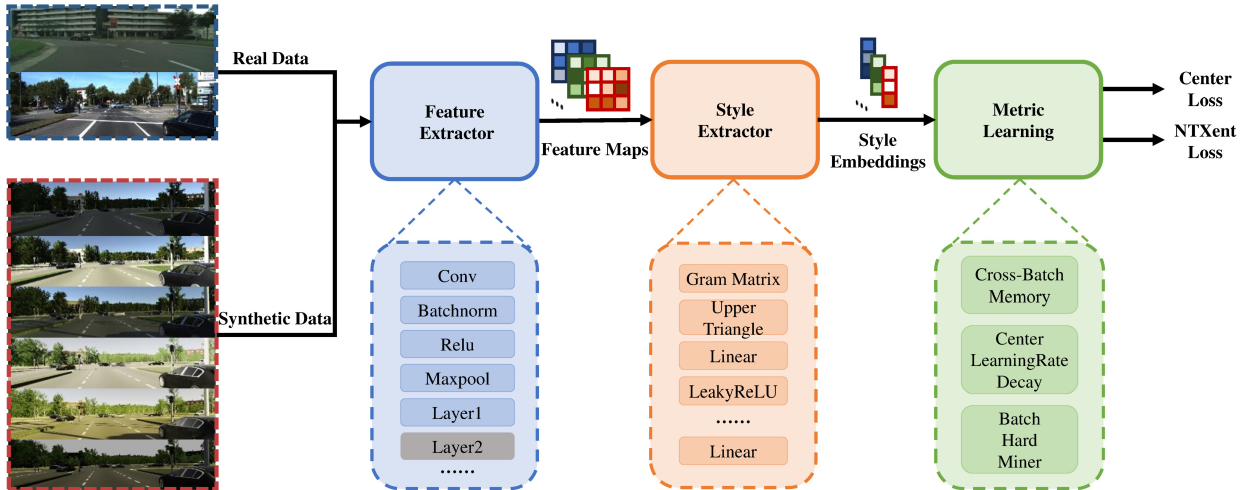} 
    \caption{The overall profiling framework. The model takes as input real images, synthetic images under various weather conditions. The input images are sequentially processed through three key modules: feature extractor, style extractor, and metric learning module. During training, the model parameters are optimized using a combination of Center Loss and NTXent Loss. In the evaluation phase, the learned feature embeddings are post-processed to compute the final SEDD metric.}
    \label{fig:lct}
    \vspace{-0.5cm}
\end{figure*}

Duminil et al. \cite{Duminil} assessed the gap by extracting texture features using the gray-level co-occurrence matrix \cite{GLCM}, local binary patterns \cite{lbp}, and discrete wavelet transform \cite{wavelet}. In their subsequent work \cite{Duminil2}, the authors introduced the Dempster-Shafer theory to enhance the evaluation model through multi-criteria fusion. Although these methods effectively reflect differences at the level of image details, their focus on texture makes them highly sensitive to environmental factors (such as weather, lighting, and scene variations). For instance, their evaluation scope is only restricted to urban scenes under clear weather conditions.

\section{Methodology}
\label{sec:method}

 The overall profiling framework is illustrated in Fig. \ref{fig:lct}. The framework takes as input real images, synthetic images under various weather conditions. During the training phase, input data is processed through a feature extractor to obtain feature maps, followed by a style extractor that generates feature embeddings. These embeddings are then processed using metric learning, incorporating Center Loss and NTXent Loss. The combined loss is propagated backward to update the parameters of the feature extractor and style extractor. In the evaluation phase, the learned feature embeddings undergo post-processing to compute fidelity metrics, providing an objective quantification of the synthetic-to-real gap. 

\subsection{Feature Extraction}
\label{subsec:featextrac}
\begin{figure}[t]
\vspace{-0.5cm} 
    \centering
    \subfloat[First layer]{%
        \includegraphics[width=0.49\linewidth]{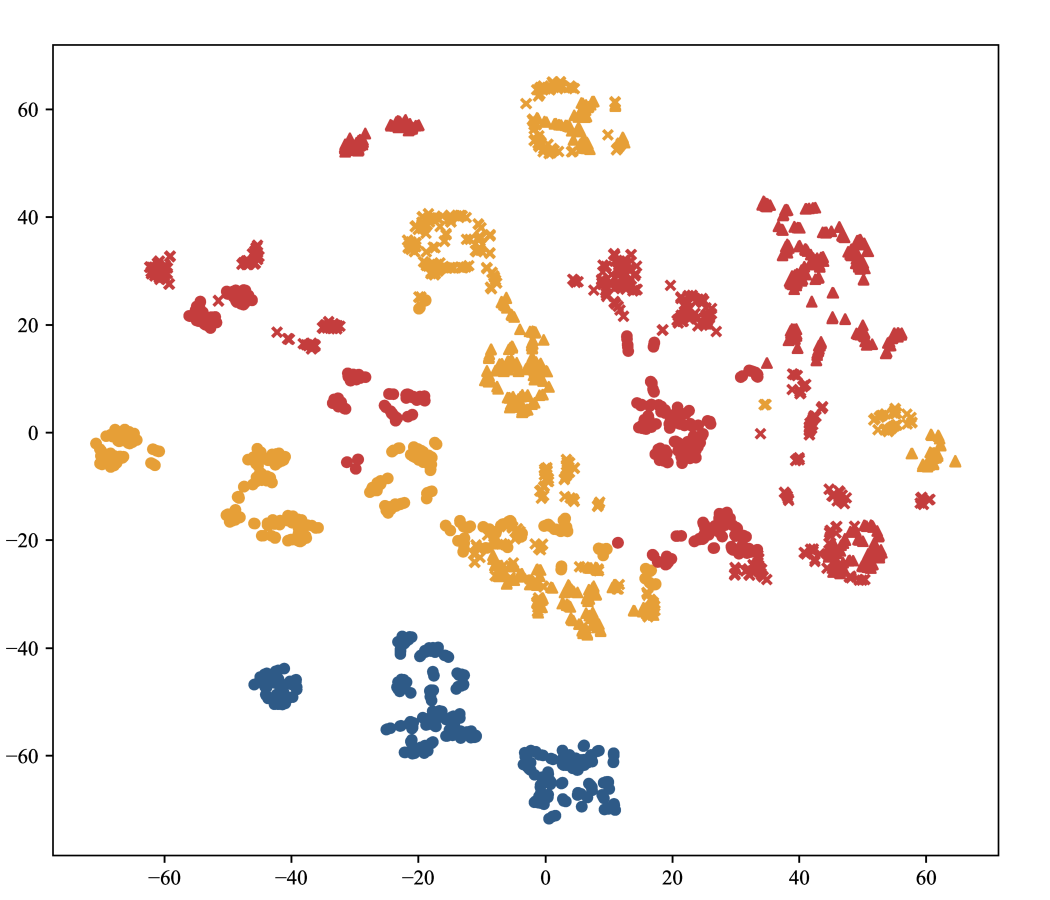}%
        \label{fig:visfeat_a}%
    }\hfill
    \subfloat[Fourth layer]{%
        \includegraphics[width=0.49\linewidth]{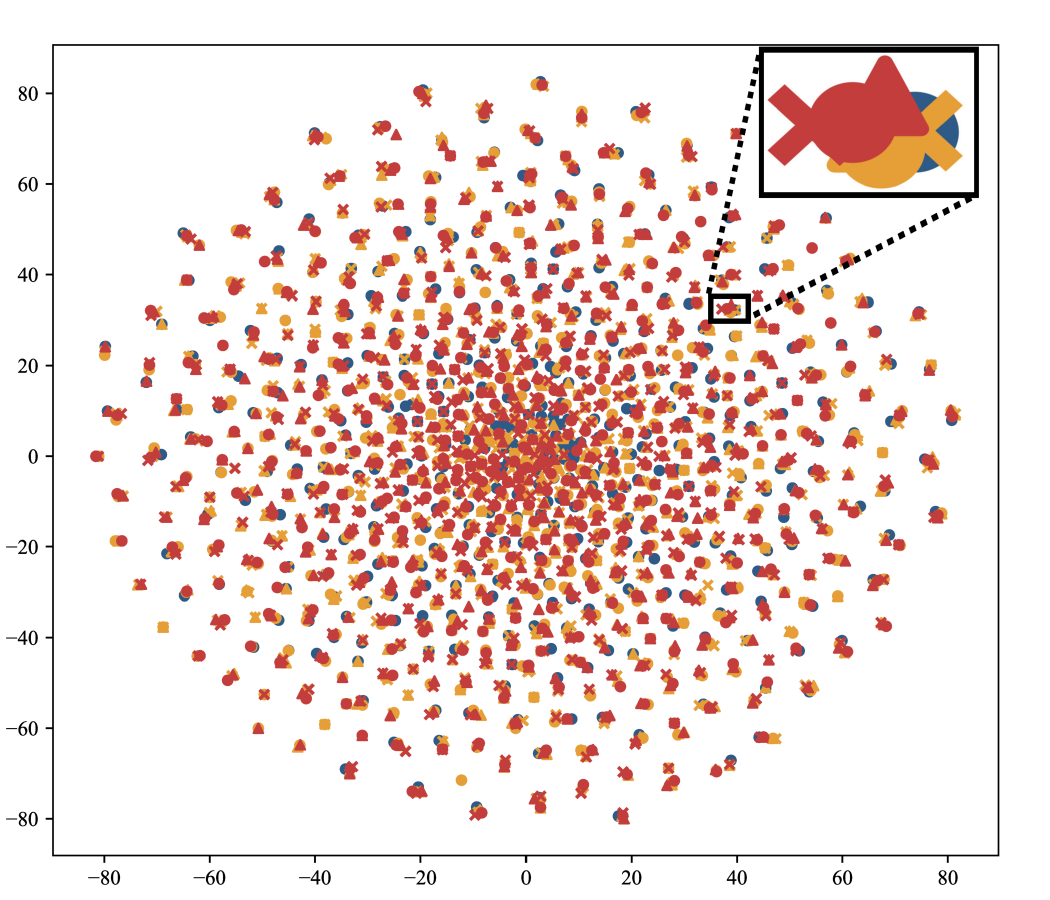}%
        \label{fig:visfeat_b}%
    }
    \caption{Visualization of feature map from different layers of ResNet. Blue points, orange points, and red points respectively represent samples from the KITTI dataset, Virtual KITTI dataset, and Virtual KITTI 2 dataset. Different shapes (circles, triangles, and crosses) denote samples under different weather conditions in the synthetic datasets.}
    \label{fig:visfeat}
    \vspace{-0.5cm}
\end{figure}

We choose ResNet \cite{resnet} as the backbone. To investigate the differences in feature representations across various layers of the feature extractor, we choose three datasets KITTI, Virtual KITTI and Virtual KITTI 2, load pre-trained weights from ImageNet \cite{ImageNet} and visualize the feature maps of different backbone layers. The visualization results are shown in Fig. \ref{fig:visfeat}. The feature maps are flattened, and then visualized using t-SNE \cite{tsne}. From the figure, it can be observed that in the shallower layers of the network, points of the same color tend to cluster together. As the depth of network increases, the points representing the same scene illustrated by circles, triangles, and crosses are observed to converge. This suggests that shallower layers primarily capture stylistic features related to dataset realism, whereas deeper layers focus more on high-level semantic information, such as scene and object content within the images. Therefore, we use the shallow layer as the feature extractor. It captures style-related features while avoiding excessive extraction of content features. 

Given an input image $I \in \mathbb{R}^{K \times H_0 \times W_0} $, the feature extractor  $\mathcal{E}$  produces the corresponding feature map  

\begin{equation}
        F = \mathcal{E}(I)
\end{equation}
where the resulting feature map \( F \in \mathbb{R}^{C \times H \times W} \), with \( C \) denoting the number of feature map channels.  

\subsection{Style Extraction}
\label{subs:styleextr}
Inspired by domain adaptation  researches \cite{fifo,divide}, we introduce Gram matrix \cite{42} for style representation. Gram matrix is computed as follows: first, each feature map is flattened into a vector of shape $C \times (H \times W)$. Then, the inner product between different channels is calculated to obtain the Gram matrix, defined by
\begin{equation}
{G}_{ij} = \mathop{\sum }\limits_{{h = 1}}^{H}\mathop{\sum }\limits_{{w = 1}}^{W}{F}_{i}\left( {h,w}\right)  \cdot  {F}_{j}\left( {h,w}\right)
  \label{eq:gram}
\end{equation}
where ${G}_{ij}$ is the Gram matrix element representing the correlation between channels $i$ and $j$.

The Gram matrix captures the similarity of response patterns between different feature maps. Instead of spatial position-dependent features, the Gram matrix focuses solely on global response patterns across the entire image, making it particularly effective for capturing spatially invariant style features. 

However, the number of elements in the Gram matrix is $C^2$, which is prohibitively large. It can have a detrimental effect on the subsequent metric learning and loss computation. To address this, we adopt a dimensionality reduction strategy.

First, since the Gram matrix $\boldsymbol{G}$ is symmetric, its upper triangular elements can fully represent the matrix. By flattening these upper triangular elements into a one-dimensional vector, we define the Gram vector as

\begin{equation}
    \boldsymbol{v} = \text{vec}(\text{triu}(\boldsymbol{G}))
\end{equation}

Then, the Gram vector is fed into a fully connected neural network, denoted as $\mathcal{S}$, where each layer consists of a linear transformation

\begin{equation}
\boldsymbol{z} = \mathcal{S}(\boldsymbol{v})
\end{equation}


This vector serves as the style embedding with dimension reduced from $C^2$ (e.g., 4096) to 64.

\subsection{Metric Learning}

The core idea of metric learning is to learn an appropriate space where samples of the same class are placed closer together, while samples of different classes are pushed farther apart.

Samples are typically divided into three categories: anchor ($A$), a reference sample; positive ($P$), from the same class as the anchor; and negative ($N$), from a different class. In our setting, all real datasets share a single label, while each synthetic dataset is treated as a separate label. For example, if an image from KITTI serves as the anchor, other real images act as positives, whereas synthetic images (e.g., from Virtual KITTI or Virtual KITTI 2) are considered negatives.

In order to efficiently store feature embeddings from historical batches, we propose the cross batch memory module. Notably, it only retains low-dimensional embeddings rather than raw images or Gram matrices. This design choice significantly reduces memory consumption while simultaneously enlarging the pool of samples available for metric learning.

To enforce intra-class compactness, we adopt Center Loss \cite{centerloss}, which minimizes the distance between a sample’s feature vector and its corresponding class center. It is defined as
\begin{equation}
    \mathcal{L}_C =\frac{1}{m} \sum_{i=1}^m \| \boldsymbol{z}_i - \boldsymbol{c}_{y_i} \|_2 
\end{equation}
where $m$ is the number of samples, $\boldsymbol{c}_{y_i}$ denotes the center of the class $y_i$, which is a learnable parameter. 

To avoid oscillations due to overly aggressive updates, we propose the center learning rate decay strategy. If the initial center learning rate is $\eta_c^0$, then at iteration $t$, the learning rate is given by
\begin{equation}
    \eta_c^{(t)} = \eta_c^0 \cdot \gamma^t 
\end{equation}
where $\gamma$ is the decay factor.

Selecting informative sample pairs is crucial for effective metric learning. The batch hard miner \cite{miner} selects challenging pairs from the current memory. For a given anchor sample $A_i$, the hard positive sample is defined as the same-class sample with the lowest similarity to $A_i$
\begin{equation}
    \mathop{\arg\min}\limits_{j: y_j = y_i} \mathop{sim}(A_i, P_j)
\end{equation}
where $\mathop{sim}$ denotes the cosine distance. Mining such difficult pairs guides the model to focus on hard negatives, which accelerates convergence and enhances overall performance.

To explicitly maximize inter-class separability, we incorporate the NTXent Loss \cite{ntxentloss}. Given an anchor sample $A_i$ with its positive counterpart $P_j$, and $S_k$ sampled from all current samples, the loss is computed as
\begin{equation}
l_{ij} = -\log \frac{\exp(\mathop{sim}(A_i,P_j)/\tau)}{\sum_{k = 1}^{2m} \mathbb{I}_{[k\neq i]}\exp(\mathop{sim}(A_i,S_k)/\tau)}
\end{equation}
\begin{equation}
  \mathcal{L}_{\text{NTXent}} = \sum_{i,j} l_{ij}  
\end{equation}
where $\tau$ is the temperature coefficient. 

\subsection{Loss Function}
To jointly optimize both intra-class compactness and inter-class separability, the overall loss function is formulated as
\begin{equation}
   \mathcal{L}_{\text{total}} = \mathcal{L}_{\text{NTXent}} + \lambda \mathcal{L}_C   
\end{equation}
where $\lambda$ is a balancing factor.

\subsection{Post-Processing}
The mean vector of style embeddings is computed as
\begin{equation}
    \boldsymbol{c} = \frac{1}{m} \sum_{i=1}^m \boldsymbol{z}_i
\end{equation}

For a new synthetic dataset, let its style center be $\boldsymbol{c}_{\text{new}}$, and let the style center of the real dataset $\boldsymbol{c}_{\text{real}}$. The Euclidean distance between these two centers is defined as $\mathit{SEDD}_1$

\begin{equation}
\mathit{SEDD}_1 = \|\boldsymbol{c}_{\text{new}} - \boldsymbol{c}_{\text{real}}\|_2 
\end{equation}

In addition, Maximum Mean Discrepancy (MMD) \cite{mmd} is a kernel-based measure for quantifying the divergence between distributions of real dataset and synthetic dataset. Using a Gaussian kernel $k(x, y) = \exp\left(-\frac{|x - y|^2}{2\sigma^2}\right)$, the squared MMD statistic between finite samples is estimated and defined as $\mathit{SEDD}_2$, calculated by
\begin{equation}
\begin{split}
    \mathit{SEDD}_2
     = &\frac{1}{m(m-1)} \sum_{i \neq j}^m  k(x_i, x_j) + \frac{1}{n(n-1)} \sum_{i \neq j}^nk(y_i, y_j)\\ &- \frac{2}{mn} \sum_{i=1}^m \sum_{j=1}^n k(x_i, y_j)
\end{split}
\end{equation}

 Euclidean distance quantifies the overall deviation of the dataset’s style, while MMD captures the distributional differences at a finer granularity.

\begin{figure*}[b]
\vspace{-0.8cm}
    \centering
    \subfloat[Results on validation set.]{%
        \includegraphics[width=0.3\linewidth]{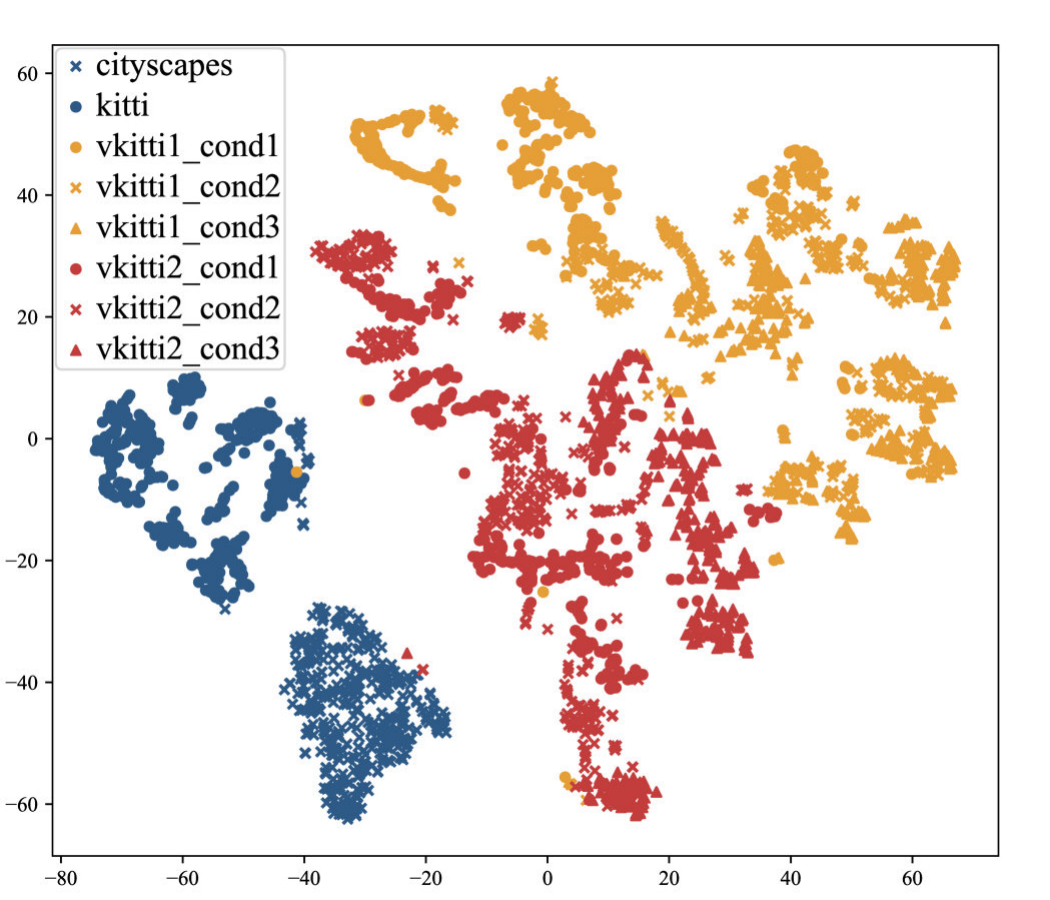}%
        \label{fig:arvs}%
    }\hfill
    \subfloat[Results on real datasets.]{%
        \includegraphics[width=0.3\linewidth]{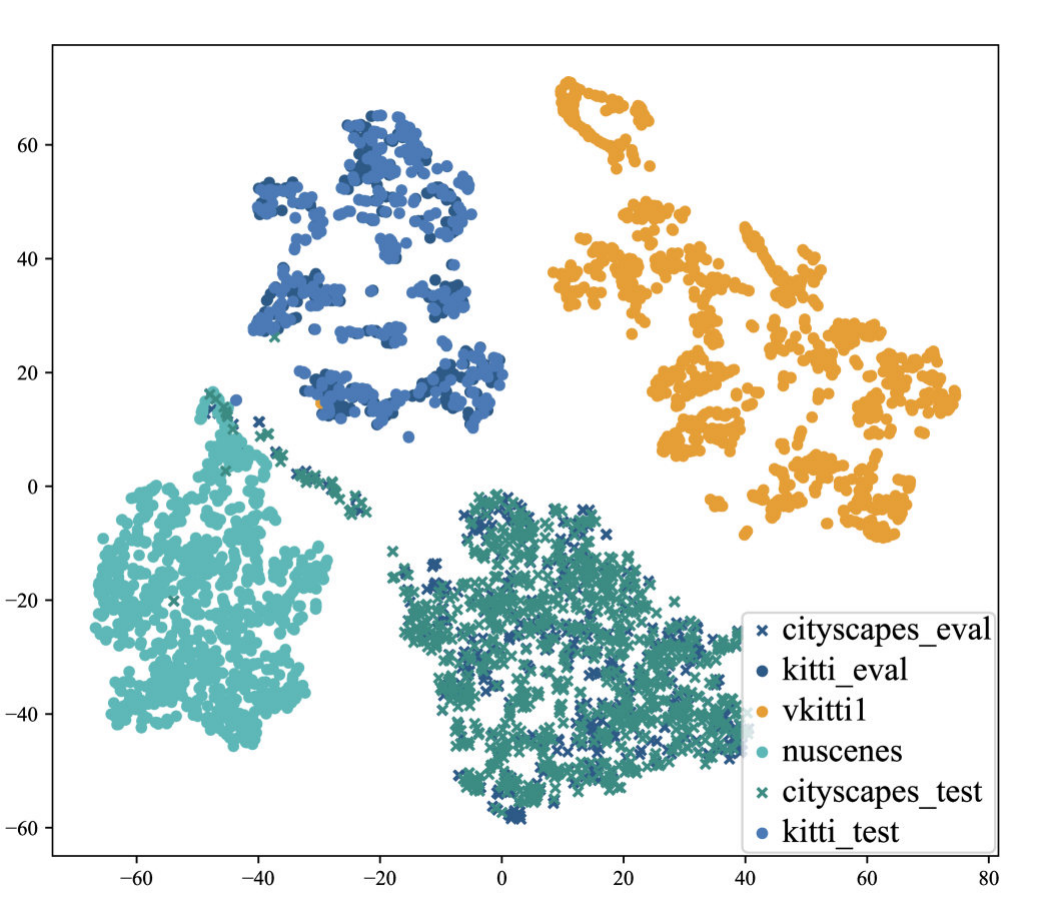}%
        \label{fig:finalreal}%
    }\hfill
    \subfloat[Results on synthetic series.]{%
        \includegraphics[width=0.3\linewidth]{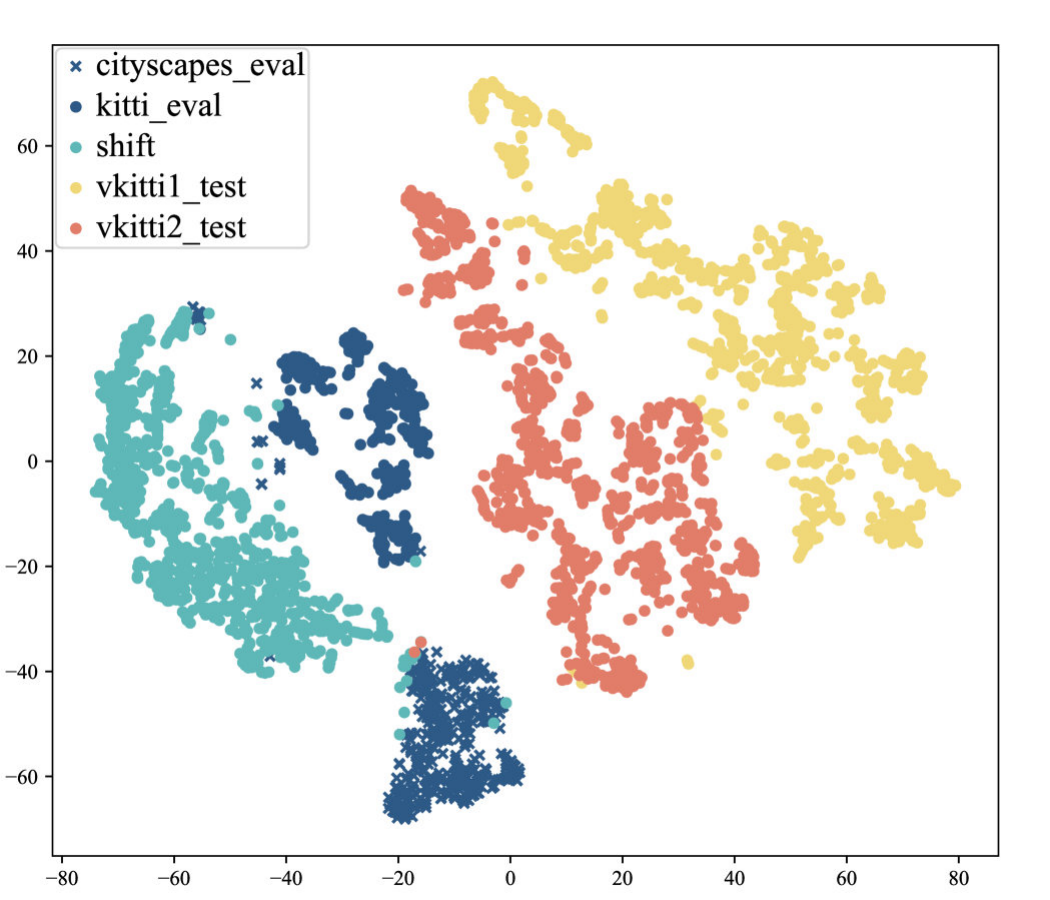}%
        \label{fig:finalsyn}%
    }\par
    \subfloat[Results for photorealism enhance series.]{%
        \includegraphics[width=0.3\linewidth]{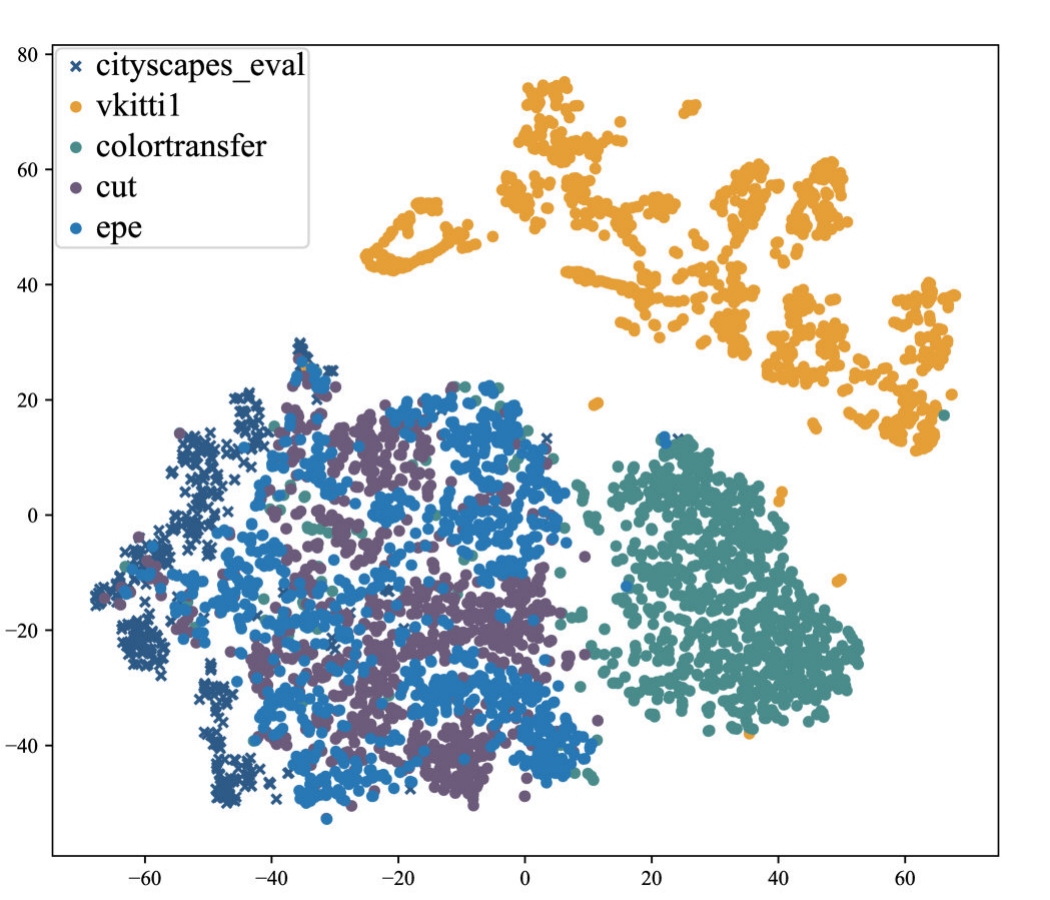}%
        \label{fig:epe}%
    }\hfill
    \subfloat[Results for CARLA2KITTI.]{%
        \includegraphics[width=0.3\linewidth]{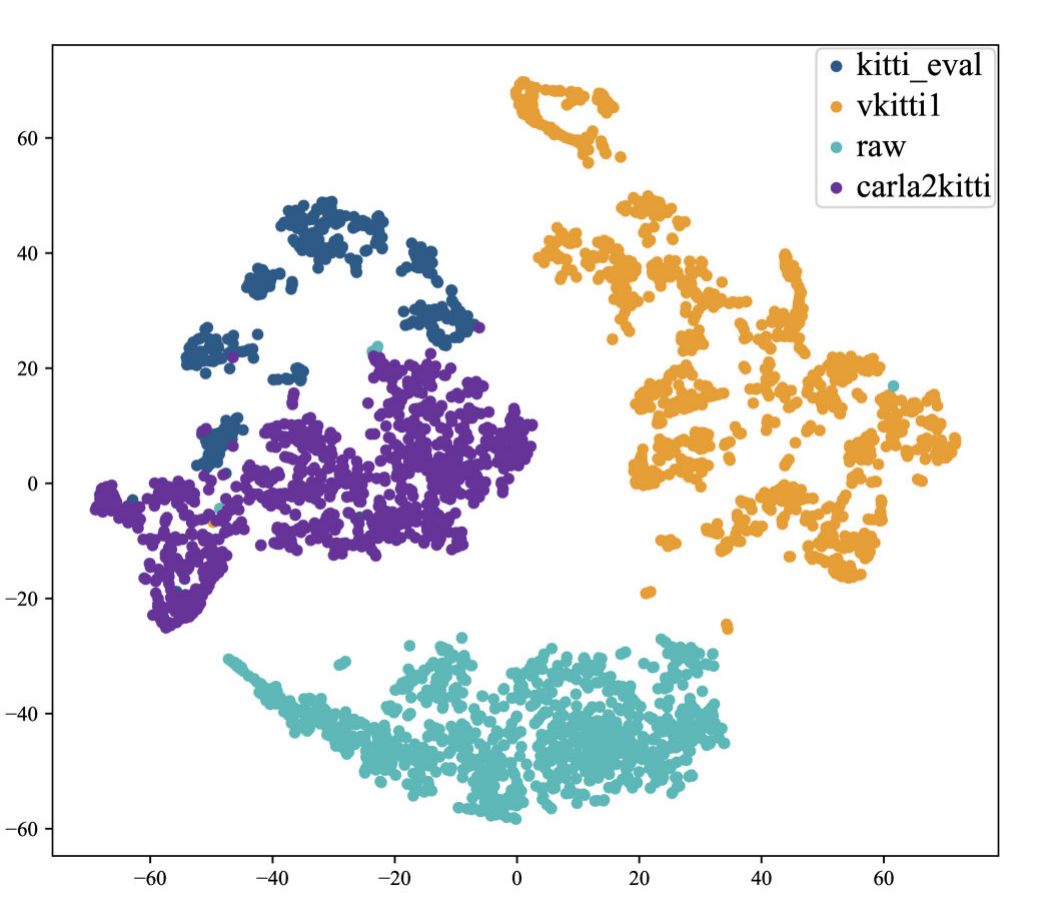}%
        \label{fig:carla2kitti}%
    }\hfill
    \subfloat[Results for CARLA2CITY.]{%
        \includegraphics[width=0.3\linewidth]{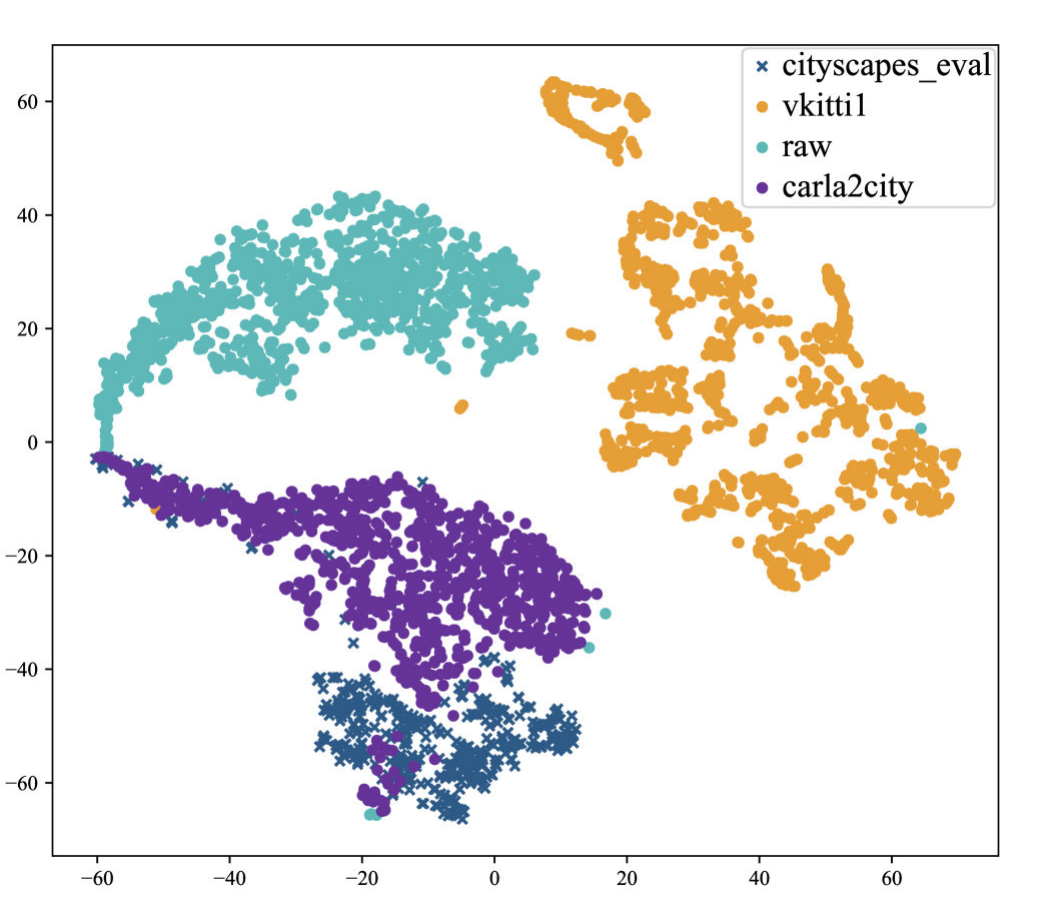}%
        \label{fig:finalcarla2real_city}%
    }
    \caption{Visualization of results on validation set and for sim-to‑real methods. The result visually demonstrates that our profiling framework can distinguish Virtual KITTI dataset from Virtual KITTI 2 dataset. And the samples after sim-to-real are closer to the real ones.}
    \label{fig:results_sim2real}
    
\end{figure*}

\section{Experiments}
\label{sec:exp}
\subsection{Experimental Setup}
\subsubsection{Datasets} Real-world datasets KITTI \cite{KITTI}, Cityscapes \cite{Cityscapes} and synthetic datasets Virtual KITTI \cite{VKITTI1}, Virtual KITTI 2 \cite{VKITTI2} are used to train our reference benchmark. For synthetic datasets, weather conditions, `clone', `morning' and `overcast', are selected. The data is partitioned into training, validation, and test sets with a ratio of 6:2:2. Additionally, we evaluate the sim-to-real methods. The photorealism enhance series include Colortransfer \cite{colortransfer}, CUT \cite{cut}, EPE \cite{EPE}. CARLA2Real \cite{carla2real} series consist of CARLA2KITTI and CARLA2CITY. For the nuScenes dataset, only the front camera images are employed. Regarding the SHIFT dataset, images set to `continuous' `10x' `front' are selected. Additionally, night-time samples are manually removed from both the nuScenes and SHIFT datasets. In training, the KITTI series of images included in each batch are specifically counterparts, i.e., the seven images are from the same moment in the same sequence.

\subsubsection{Implementation Details} All experiments are conducted on a single NVIDIA GeForce RTX 3090 GPU. ResNet-18 \cite{resnet} is selected for feature extractor backbone. The feature extractor is initialized with pre-trained weights from ImageNet \cite{ImageNet}. The cross batch memory size is set to 100 and kernel bandwidth $\sigma$ is set to 10. The model is trained for a maximum of 4 epochs, with a learning rate of 5e-5 for the feature extractor and style extractor. The hyperparameter $\lambda $ is set to 0.5 and $\tau $ is set to 0.015. We use the SGD optimizer with momentum of 0.9 and weight decay of 1e-4. The initial learning rate of Center Loss is set to 10, and the decay factor $\gamma$ is set to 0.9. 

\subsubsection{Visualization} We employ t-SNE \cite{tsne} for visualization. For visual effect, 1000 images were randomly selected from each test dataset. It is worth noting that due to the visualization algorithm itself, even for the same data distribution, its shape may change after t-SNE with other different distributions.

\subsection{Results and Discussions}

\subsubsection{Quantification Results on Validation Set} The results of our trained profiling framework on the validation set are shown in Table \ref{tab:stand}. This result provides a reference value for subsequent quantification. Notably, during the training process, no label prior information indicating that Virtual KITTI 2 is superior was provided. Nevertheless, our profiling framework autonomously determines that the style of Virtual KITTI 2 is closer to that of the real dataset, resulting in a smaller synthetic-to-real gap. Fig. \ref{fig:arvs} presents the visualization results. In the figure, the sample points are clustered by color rather than by shape, which indicates that our method achieves the disentanglement of content and style. 


\begin{table}[htbp]
\begin{center}
    \caption{Quantification results on validation set}
    \label{tab:stand}
  \begin{tabular}{@{}ccc@{}}
    \toprule
     Datasets  & $\mathit{SEDD}_1$ &  $\mathit{SEDD}_2$ \\
    \midrule
    Virtual KITTI &   0.308  & 0.488 \\
   Virtual KITTI 2 &  0.276  &  0.408 \\
    \bottomrule
  \end{tabular}
  \end{center}
\end{table}

\vspace{-0.5cm} 

\begin{table}[ht]
    \centering
\caption{Comparison with NR-IQA methods on test set}
  \label{tab:iqa}
  \begin{tabular}{@{}cccccc@{}}
    \toprule
Datasets & \thead{Brisque \\ $\downarrow$ \cite{brisque}} & \thead{NIMA \\ $\uparrow$ \cite{nima}}  & \thead{Unique \\ $\uparrow$ \cite{unique}}  & $\mathit{SEDD}_1$ $\downarrow$ & $\mathit{SEDD}_2$ $\downarrow$ \\
    \midrule
     \textbf{KITTI} & \underline{21.6} & 3.73&0.11 &\textbf{0.102} & \textbf{0.081} \\
    VKITTI &42.7 &\textbf{4.56} & \underline{0.13}&0.305 &0.480 \\
   \underline{VKITTI2} & \textbf{19.1} &\textbf{4.56} &  \textbf{1.40}   & \underline{0.274} & \underline{0.402} \\
    \bottomrule
  \end{tabular}
\end{table}

\subsubsection{Compared with NR-IQA} The results are summarized in Table \ref{tab:iqa}, where the best performance is highlighted in bold and the second-best is underlined. The results demonstrate that NR-IQA methods do not sort correctly, while only our method correctly identifies KITTI as the most realistic and Virtual KITTI as the least realistic. This comparison highlights a fundamental advantage of our approach: unlike NR-IQA metrics that prioritize perceptual image quality, often erroneously assigning higher scores to clear but unrealistic synthetic images, SEDD explicitly measures the distributional deviation from the real domain style. This ensures a fidelity ranking that aligns with objective realism rather than mere visual clarity.

\begin{table*}[t]
\centering
\caption{Experiments on real and synthetic test set  }
  \label{tab:realandsyn}
  
\begin{tabular}{c|c|c|cccc|cc}
\toprule
\multirow{2}{*}{Datasets} & \multirow{2}{*}{FE$\downarrow$\cite{gadipudi}} & \multirow{2}{*}{Li et al. $\uparrow^5$\cite{Li}} & \multicolumn{4}{c|}{Duminil et al. $\uparrow^{100}$\cite{Duminil,Duminil2}} & \multicolumn{2}{c}{Ours$\downarrow$} \\
\cline{4-9}
 & & & GLCM & Wavelets & LBP & sH & \multicolumn{1}{c}{$\mathit{SEDD}_1$} & \multicolumn{1}{c}{$\mathit{SEDD}_2$} \\
\hline
KITTI\cite{KITTI} & -   &4.845 & 99.81 & 96.85 & 94.87 & 62.42 &\textbf{0.103}  &\textbf{0.085}  \\
Cityscapes\cite{Cityscapes} & 4.47 & - & 96.67 &98.05  & 87.46 & 72.76 &\textbf{0.122} & \textbf{0.111} \\
nuScenes\cite{nuScenes} & -& -  & 90.94 & 98.54 & 60.46  & 73.60 & \textbf{0.143} &\textbf{0.197}  \\
\hline
Virtual KITTI\cite{VKITTI1} &22.31 & -  & - & - &-  & - &\textbf{0.316}  & \textbf{0.494} \\
Virtual KITTI 2\cite{VKITTI2} &- & - &0.05   &51.21  &02.24   &34.10  & \textbf{0.275} &  \textbf{0.406} \\
SHIFT\cite{shift} &- & 4.792 & - & - & - & - & \textbf{0.210} & \textbf{0.275} \\
\bottomrule
\end{tabular}
\end{table*}

\subsubsection{Quantification Results on Test Set} The results of our model on test set  are shown in Table \ref{tab:realandsyn}. `-' indicates that the experiment was not performed. This absence stems from the lack of open-source implementations for certain baselines, which prevented a complete comparative evaluation. The  $\mathit{SEDD}_1$ and $\mathit{SEDD}_2$ of the real dataset are all smaller than those of the synthetic dataset. The visualization results on the real datasets are shown in Fig. \ref{fig:finalreal} and results on the synthetic datasets are shown in Fig. \ref{fig:finalsyn}. Unlike other methods that have not been tested on both Virtual KITTI and Virtual KITTI 2, two datasets with extremely similar scenarios, our approach is capable of making accurate distinctions and aligns well with objective realism. Furthermore, unlike our framework, these prior methods were not originally designed to account for generalization across unseen domains. For instance, \cite{gadipudi} fails to disentangle content from style, making them susceptible to content variations. Similarly, \cite{Duminil} and \cite{Duminil2} are highly sensitive to environmental factors such as weather and lighting, restricting their scope primarily to clear urban scenes. In contrast, our work explicitly addresses these limitations by validating the framework on the unseen nuScenes and SHIFT datasets. 

\begin{table}[htbp]
    \centering
    \caption{Experiments for sim-to-real methods }
  \label{tab:sim2real}
\begin{tabular}{l|l|cc}
    \toprule
    \multicolumn{2}{c|}{Sim-to-real Methods} & $\mathit{SEDD}_1$ & $\mathit{SEDD}_2$ \\
    \cline{1-4}
    \multirow{4}{*}{\makecell{Photo-\\realism\\Enhance}} 
    & Colortransfer\cite{colortransfer} & 0.242 &0.384  \\
    & CUT\cite{cut} & 0.192  & 0.278\\
    & EPE2Vistas\cite{EPE,vistas}   &  0.191 &0.249  \\
    & EPE\cite{EPE}   &  \textbf{0.175}   &\textbf{0.245}  \\
    \hline
    \multirow{3}{*}{\makecell{CARLA\\2Real}} 
    & Raw(CARLA)\cite{carla}          &0.193  & 0.307\\
    & CARLA2KITTI\cite{carla2real}  &  \textbf{0.172}                                 & \textbf{0.158} \\
    & CARLA2CITY\cite{carla2real}   &  \textbf{0.135}                                 &\textbf{0.170}  \\

    \bottomrule
\end{tabular}

\end{table}

\subsubsection{Sim-to-real Gap Quantification} The results for sim-to-real methods are presented in Table \ref{tab:sim2real}. The smaller the value, the more effective the sim-to-real method. Visualization results for photorealism enhance series are shown in Fig. \ref{fig:epe} and for CARLA2Real series are shown in Fig. \ref{fig:carla2kitti} and Fig. \ref{fig:finalcarla2real_city}. The enhanced images exhibit lower $\mathit{SEDD}_1$ and $\mathit{SEDD}_2$ values, indicating that our proposed method effectively quantifies the impact of sim-to-real translation. By providing a continuous quantitative measure, our metric allows researchers to precisely compare the effectiveness of different adaptation strategies (e.g., differentiating the superior performance of EPE over CUT), thereby serving as a definitive benchmark for guiding algorithm selection.

\subsubsection{Generalization}
\label{sec:general}
Experiments are conducted on real dataset nuScenes \cite{nuScenes}, synthetic dataset SHIFT \cite{shift} and sim-to-real method EPE2Vistas \cite{EPE} (refers to the EPE method where the target dataset is Vistas \cite{vistas}). None of these datasets appears in the training process, but their performances meet the expectations, indicating our generalization ability.

\subsection{Ablation Study}

\begin{wraptable}{r}{3.5cm}
    \centering
      \caption{Ablation study.}
  \label{tab:ablation}
  \begin{tabular}{@{}lc@{}}
    \toprule
     Settings  & Variance \\
    \midrule
   only $\mathcal{L}_{\text{NTXent}}$ &  0.250   \\
   only $  \mathcal{L}_C  $  &   0.230  \\
    $    \mathcal{L}_{\text{total}}$ &\textbf{ 0.224 }    \\
    \bottomrule
  \end{tabular}
  \vspace{-0.3cm} 
\end{wraptable}

 We evaluated the effect of the loss function using the internal variance of the real dataset. As shown in Table \ref{tab:ablation}, the experimental results indicate that using a single loss function leads to an increase in the intra-class variance of the feature distributions, whereas the joint loss effectively constrains the compactness of intra-class samples. The ablation experiments further demonstrate that solely relying on $\mathcal{L}_{\text{NTXent}}$ may cause the framework to overlook intra-class consistency, highlighting the need for an explicit intra-class constraint. In contrast, optimizing \(\mathcal{L}_C\) in isolation tends to converge to a local optimum, while the inter-class information provided by the contrastive loss helps mitigate it.


\subsection{Hyperparameter Experiment}
\label{sec:hyper}
Table \ref{tab:hyper} presents the results of varying the temperature $\tau$ and the loss function balancing factor $\lambda$.  The results are expressed as `real dataset variance / center distance between synthetic and real datasets'. The degree of intra-class aggregation is represented by the variance of real dataset, where a smaller value is preferable. The degree of inter-class separation is measured by the distance between the centers of the Virtual KITTI and the real dataset. The final model is a trade-off between the intra-class aggregation and inter-class separation.

\begin{table}[htbp]
  \centering
    \caption{Hyperparameter experiment}
  \label{tab:hyper}
  \begin{tabular}{c|ccc}
    \toprule
    \diagbox[height=1.5em]{$\tau$}{$\lambda$} & 0.2 &\textbf{0.5} &1.0 \\
    \hline
    0.010  & 0.250 / 0.363 & 0.245 / 0.361& 0.256 / 0.391\\
    \textbf{0.015}  &0.228 / 0.306  &  \textbf{0.224 / 0.304} & 0.228 / 0.318\\
     0.030  &0.217 / 0.263  & 0.214 / 0.264& 0.215 / 0.271\\
    \bottomrule
  \end{tabular}

\end{table}

\subsection{Discussion}
\subsubsection{Hyperparameter and Trade-offs}
The sensitivity analysis of the temperature $\tau$ and the balancing factor $\lambda$ presented in \ref{sec:hyper} serves to characterize the model’s behavior rather than to locate a singular global optimum. This is because the evaluation indicators, intra-class aggregation and inter-class separation, are inherently adversarial. As illustrated in Table \ref{tab:hyper}, varying $\tau$ and $\lambda$ results in a predictable shift where an improvement in one metric often leads to a slight degradation in the other. However, the overall framework effectively converges to a stable equilibrium. Consequently, the grid search conducted only on the Virtual KITTI benchmark is sufficient. Since the primary goal is to identify a stable operating range where both constraints are reasonably satisfied, rather than to overfit the parameters to maximize a specific metric, fine-grained tuning or exhaustive search is unnecessary. This indicates that the framework is not critically sensitive to precise parameter selection, and the established equilibrium is robust enough to generalize without dataset-specific re-optimization.
\subsubsection{Limitations}
Despite the strong generalization capabilities demonstrated on unseen datasets such as nuScenes, SHIFT and EPE2Vistas, as well as across distinct weather conditions, we acknowledge that the framework may still encounter limitations when applied to scenarios characterized by extremely high diversity or complexity. For instance, samples with extreme illumination conditions, such as night-time scenes. In such cases, the capacity of the current feature and style extraction modules to fully disentangle style from content may be challenged. Future research could investigate more advanced architectures for feature and style extraction to achieve a more rigorous disentanglement.
\section{Conclusion}
\label{sec:con}

In this work, we propose a novel style-based profiling framework  with the SEDD metric. Our approach leverages ResNet-based architectures as the backbone network for extracting effective features from input images. To capture the stylistic characteristics, we employ Gram matrices to generate style embeddings, which are subsequently processed through fully connected neural networks for dimensionality reduction while maintaining essential style information. The proposed metric learning framework utilizes Center Loss and NTXent Loss to optimize the model. Extensive experiments validate the effectiveness of the proposed metric. 

This work transforms the synthetic-to-real gap into quantifiable objective indicators, eliminating reliance on subjective experience. It provides a standardized proactive tool for the quality control of synthetic datasets in autonomous driving, facilitating more targeted improvements. Future work could explore leveraging SEDD as an objective function for generative models, creating a closed-loop system that actively guides high-fidelity data synthesis.
	
	\bibliographystyle{IEEEtran}
	\bibliography{root} 
	
\end{document}